\documentclass[letterpaper]{article}
\usepackage{myaaai08}
\usepackage{times}
\usepackage{helvet}
\usepackage{courier}

\usepackage{boxedminipage}
\usepackage{graphicx}
\usepackage{epsfig}
\usepackage{amssymb}
\usepackage{amsmath}
\usepackage{amsfonts}
\usepackage{subfigure}
\newcommand{\fs}[1]{\fontsize{#1}{#1}\selectfont}

\newtheorem{definition}{Definition}  
 
\newtheorem{thm}{Theorem}

\def\lph{$\mathcal{LPH}$}
\def\htnpref{\textbf{\textsc{htnpref}}}
\def\final{\textbf{final}} 
\def\nextt{\textbf{next}} 
 
\def\always{\textbf{always}} 
\def\eventually{\textbf{eventually}} 
\def\until{\textbf{until}}

\def\true{\mbox{{\small TRUE}}}
\def\false{\mbox{{\small FALSE}}}
\newcommand{\pr}{{\rho_{s}}} 
\newcommand{\minval}{{\small v_{min}}}
\newcommand{\maxval}{{\small v_{max}}}
\newcommand{\before}{{\textbf{before}}} 
\newcommand{\holdBefore}{{\textbf{holdBefore}}} 
\newcommand{\holdAfter}{{\textbf{holdAfter}}} 
\newcommand{\holdBetween}{{\textbf{holdBetween}}}

\newcommand{\occ}{{\textbf{occ}}} 
\newcommand{\apply}{{\textbf{apply}}}

\newcommand{\occnext}{{\textbf{occNext}}}

\newcommand{\sSo}{{\small $S_0$}}

\newcommand{\ssit}{{\small $s$}}

\newcommand{\sdas}{{\small $do(a,s)$}}
\def\atheory{\mathcal{D}}

\newcommand{\isdef}{\hbox{~$\stackrel{\rm def}{=}$~}}

\def\terminated{\textit{terminated}}
\def\executing{\textit{executing}}
\def\start{\textit{start}}
\def\endd{\textit{end}}

\def\badSituation{\textit{badSituation}}

\def\expand{\textbf{\textsc{expand}}}

\def\initfront{\textbf{\textsc{initFrontier}}}
\def\remfirst{\textbf{\textsc{removeFirst}}}
\def\sortnmerge{\textbf{\textsc{sortNmerge}}}

\def\aspen{\textbf{\textsc{aspen}}}

\def\shop2{\textbf{\textsc{shop2}}}
\def\NC{\textbf{\textsc{NC}}}
\def\NE{\textbf{\textsc{NE}}}
\def\PL{\textbf{\textsc{PL}}}

\def\enquirer{\textbf{\textsc{enquirer}}}

\def\occ{\textbf{occ}} 
\def\apply{\textbf{apply}} 
\def\applynext{\textbf{applyNext}}
 
\def\occnext{\textbf{occNext}}
\def\final{\textbf{final}} 
\def\nextt{\textbf{next}} 
 
\def\always{\textbf{always}} 
\def\eventually{\textbf{eventually}} 
\def\until{\textbf{until}}

\def\gand{\, \& \,} 
\def\gor{\,\mid \,}

\def\min{\mbox{min }} 
\def\pr{\rho_{s}}

\def\vset{\mathcal{V}} 
\def\minval{v_{min}}
\def\maxval{v_{max}}

\def\atheory{\mathcal{D}}

\def\lpp{$\mathcal{LPP}$}

\def\expand{\textbf{\textsc{expand}}}

\def\tab{\hspace*{.25in}} 
 
\def\bigtab{\hspace*{.4in}}  
\renewcommand{\theequation}{P\arabic{equation}}

\begin{document}

\title{On Planning with Preferences in HTN}

\author{Shirin Sohrabi \and Sheila A. McIlraith \\
Department of Computer Science \\
University of Toronto \\ 
Toronto, Canada. \\
\{shirin,sheila\}@cs.toronto.edu 
}

\nocopyright

\maketitle

\begin{abstract}

In this paper, we address the problem of generating preferred plans
by combining the procedural control knowledge specified
by Hierarchical Task Networks (HTNs) with rich qualitative
user preferences. The outcome of our work is a language for specifying
user preferences, tailored to HTN planning, together with a provably
optimal preference-based planner, \htnpref, that is implemented as
an extension of \shop2. To compute preferred plans,
we propose an approach based on forward-chaining heuristic search. Our heuristic uses an admissible evaluation function measuring
the satisfaction of preferences over partial plans. 
Our empirical evaluation demonstrates the effectiveness of
our \htnpref~heuristics. We prove our 
approach sound and optimal with respect to the plans it generates 
by appealing to a situation calculus semantics of our preference
language and of HTN planning. While our implementation builds on
\shop2, the language and techniques proposed here are relevant to a
broad range of HTN planners.

\end{abstract}

\section{Introduction}
\label{section-introduction}

Hierarchical Task Network (HTN) planning is a popular and widely used 
planning paradigm, and many domain-independent HTN
planners exist (e.g., \shop2, {\small SIPE-2, I-X/I-PLAN, O-PLAN}) \cite{htn04}.
In HTN planning, the planner is provided with
a set of tasks to be performed, possibly together with constraints on 
those tasks.  A plan is then formulated by repeatedly decomposing tasks 
into smaller and smaller subtasks until primitive, executable tasks
are reached.  
A primary reason behind HTN's success is that its task networks capture
useful procedural control knowledge---advice on how to
perform a task---described in terms of a decomposition of subtasks.
Such control knowledge can significantly reduce the search space for a plan while also
ensuring that plans follow one of the stipulated courses of action.
However, while HTNs specify a family of satisfactory
plans, they are, for the most part, unable to distinguish what constitutes
a high-quality plan.

In this paper, we address the problem of generating preferred
plans by augmenting HTN planning problems with rich qualitative
user preferences.  User preferences can be arbitrarily complex, 
often involving 
combinations of conditional, interacting, and mutually exclusive 
preferences that can range over multiple states of a plan.  This
makes finding an optimal plan hard.  There are two aspects to 
addressing the problem of preference-based planning with HTNs.  
The first is to propose a preference specification language that 
is tailored to HTN planning.  The second, is to generate preferred, 
and ideally optimal, plans efficiently.  

To specify user preferences, we augment a rich qualitative 
preference language, \lpp, proposed in \cite{biefrimcikr06} with HTN-specific
constructs.  \lpp~specifies 
preferences in a variant of linear temporal logic (LTL).  Among the 
HTN-specific properties that we add to our language, \lph, is the
ability to express preferences over how tasks in our HTN are decomposed
into subtasks, preferences over the parameterizations of decomposed
tasks, and a variety of temporal and nontemporal preferences over the
task networks themselves.   

To compute preferred plans, we propose an approach based on 
forward-chaining heuristic search.  Key to our approach is a
means of evaluating the (partial) satisfaction of preferences
during HTN plan generation based on progression.
The optimistic evaluation of  
preferences yields an admissible evaluation function which we use 
to guide search.   
We implemented our planner, \htnpref, 
as an extension to the \shop2~HTN planner.
Our empirical evaluation demonstrates the effectiveness of \htnpref~heuristics in finding high-quality plans. 
We provide a semantics for our preference language  
in the situation calculus \cite{reiter01} and appeal to this semantics 
to prove the soundness and optimality of our planner with respect to the 
plans it generates.  This paper omits a number of technical details
that can be found in a longer paper describing this work.

\section{HTN Planning}
\label{htn-background}

In this section, we provide a brief overview of both HTN planning,
following \cite{htn04}, and our situation calculus encoding
of preference-based HTN planning.

\noindent {\bf Travel Example:} 
Consider a simple HTN planning problem to address the task of
arranging travel.  This task can be decomposed into arranging transportation,
accommodations, and local transportation.  Each of these tasks can
again be decomposed based on alternate modes of transportation and
accommodations, reducing eventually to primitive actions that can
be executed in the world.  Further constraints can be imposed
to restrict decompositions.

\begin{small}
\begin{definition}[HTN Planning Problem] An HTN planning problem is a 
3-tuple $\mathcal{P} = (s_{0}, w, D)$ where $s_{0}$ is the initial state, $w$ 
is a task network called the \textit{initial task network}, and $D$
is the HTN planning domain.
$\mathcal{P}$ is a total-order planning problem if w 
and $D$ are totally ordered; otherwise it is said to be partially ordered.
\end{definition}
\end{small}

A \emph{task} consists of a task symbol and a list of arguments.
A task is primitive if its task symbol is an operator name and
its parameters match, otherwise it is \emph{nonprimitive}. 
In our example, \textit{arrange-trans} and \textit{arrange-acc} are
nonprimitive tasks, while \textit{book-flight} and \textit{book-car} are
primitive tasks.

\begin{small}\begin{definition}[Task Network] A task network is a pair w=(U, C)
where U is a set of task nodes and C is a set of constraints.
Each task node u $\in$ U contains a task $t_{u}$. If all of the tasks
are ground then w is ground; 
If all of the tasks
are primitive, then w is called primitive; otherwise is called nonprimitive.
Task network $w$ is \textit{totally ordered} if $C$ defines
a total ordering of the nodes in U. 
\end{definition}\end{small}

In our example, we could have a task network $(U, C)$ where $U=\{u_{1}, u_{2} \}$, $u_{1}=$\textit{book-car}, and $u_{2}$= \textit{pay}, and $C$ is a precedence constraint such that $u_{1}$ must occur before $u_{2}$ and a before-constraint  such that at least one car is available for rent before $u_{1}$. 

A domain 
is a pair {\small $D = (O, M)$} 
where {\small $O$} is a set of operators and {\small $M$} is a set of methods. Operators are essentially 
primitive actions that can be executed in the world.  They
are described by a triple  \textit{o =(name(o), pre(o), eff(o))},
corresponding to the operator's name, preconditions and effects.
Preconditions are restricted to a set of literals, and effects are
described as STRIPS-like Add and Delete lists.  
An operator $o$ can accomplish a ground
primitive task in a state $s$ if their names match and $o$ is applicable
in $s$.
In our example, ignoring the parameters,
operators might include:  \textit{pay,  book-train, 
book-car, book-hotel,} and \textit{book-flight}.

A method, {\small $m$}, is a 4-tuple (\textit{name(m), task(m),subtasks(m), 
constr(m))} corresponding to the method's name, a
nonprimitive task and the method's task network,
comprising subtasks and constraints. 
A method is \textit{totally ordered} if its task network is \textit{totally ordered}.
A domain is a total-order domain if every $m \in M$ is \textit{totally ordered}.
Method $m$ is relevant for a task $t$ if there is a substitution 
$\sigma$ such that  $\sigma(t)$ =\textit{task}($m$). 
Several different methods can be relevant to a particular nonprimitive
task $t$, leading to different decompositions of $t$.
In our example, the method with  \textit{name} \textit{by-flight-trans} can
be used to decompose the \textit{task} \textit{arrange-trans}
into the \textit{subtasks} of booking a flight and paying, 
with the constraint (\textit{constr}) that the booking precede payment.


\begin{small}
\begin{definition}[Solution to HTN Planning Problem] Given HTN planning problem $\mathcal{P} = (s_{0}, w, D)$, 
a plan $\pi=(o_{1}, ..., o_{k})$ is a solution for $\mathcal{P}$, 
depending on these two cases:
1) if w is primitive, then there must exist a ground instance of 
($U', C'$) of (U, C) and a total ordering $(u_{1}, ..., u_{k})$ of the nodes in $U'$ such that for all $1 \leq i \leq k$, name($o_{i}$) = $t_{u_{i}}$, the plan $\pi$ is executable in the state $s_{0}$, and all the constrains  hold, 2) if w is nonprimitive, then there must exist a sequence of 
task decompositions that can be applied to w to produce a 
primitive task network $w'$, where $\pi$ is a solution for $w'$. 
\end{definition}
\end{small}

Finally, we define the HTN preference-based planning problem.
This definition appeals to two concepts that are not yet well-defined
and which we defer to later sections:  
definitions of the form and content of the
the formula $\Phi_{htn}$ that captures user preferences for HTN planning as well as 
and the precise definition of 
\emph{more preferred} appears in Section 3.

\begin{small}
\begin{definition}[Preference-based HTN Planning] 
\label{def:pbhtnp}
An HTN planning problem with user preferences is described as a 4-tuple
$\mathcal{P}= (s_{0}, w, D, \Phi_{htn})$ where 
$\Phi_{htn}$ is a formula describing user preferences. 
A plan $\pi$ is a solution to $\mathcal{P}$ if and only if: 
$\pi$ is a plan for $\mathcal{P'}= (s_{0}, w, D)$ and there 
does not exists a plan $\pi'$ such that $\pi'$ is more 
preferred than $\pi$ with respect to the preference formula $\Phi_{htn}$.
\end{definition}
\end{small}

\subsection{Situation Calculus Specification of HTN}

We now have a definition of preference-based HTN planning.  Later in the
paper, we propose an approach to computing preferred plans, together
with a description of our implementation.  To prove
the correctness and optimality of our algorithm, we appeal to an existing
situation calculus encoding of HTN planning, which we augment and
extend to provide an encoding of preference-based HTN planning. Since
the situation calculus has a well-defined semantics, we have a
semantics for our encoding which we use in our proofs.  
In this section, we review the salient features of this encoding.

{\bf The Situation Calculus} is a logical language for 
specifying and reasoning about dynamical systems \cite{reiter01}. 
In the situation calculus, the {\em state} of the world is expressed in 
terms of functions and relations (fluents) relativized to a particular 
{\em situation} \ssit, e.g., {\small $F(\vec{x},s)$}. 
A situation \ssit ~is a {\em history} of the primitive actions, 
{\small $a \in {\cal A}$}, performed from a distinguished initial
situation \sSo. The function \sdas~maps a situation and an action 
into a new situation thus inducing a tree of situations rooted in \sSo. 
A \textit{basic action theory} in the situation calculus $\atheory$ 
includes \textit{domain independent foundational axioms}, 
and \textit{domain dependent axioms}.
A situation $s'$ precedes a situation $s$, i.e., $s' \sqsubset s$, 
means that the sequence $s'$ is a proper prefix of sequence $s$.

{\bf Golog}
\cite{reiter01} is a high-level logic programming
language for the specification and execution of complex actions
in dynamical domains. It builds on top of the situation calculus
by providing Algol-inspired extralogical constructs for
assembling primitive situation calculus actions into complex
actions  ({\em programs}) {\small $\delta$}.
Example complex actions include
action sequences, if-then-else, while loops, nondeterministic choice
of actions and action arguments, and procedures. These complex
actions serve as constraints upon the situation tree. 
ConGolog \cite{giacomo00} is the concurrent version of Golog in 
which the language can additionally deal with execution of concurrent processes, interrupts, prioritized concurrency, 
and exogenous actions.

A number of researchers have pointed out the connection between HTN and 
ConGolog. 
Following Gabaldon \cite{alfredohtn02}, 
we map an HTN state to a situation calculus \textit{situation}.
Consequently, the initial HTN state $s_{0}$ is encoded as the
initial situation, {\small $S_{0}$}. The HTN domain description maps to a 
corresponding situation calculus domain description, $\atheory$, where
for every operator $o$ there is a corresponding primitive action 
$a$, such that the preconditions and the effects 
of $o$ are axiomatized in $\atheory$.
Every method and nonprimitive task together with constraints is encoded as a ConGolog procedure.
For the purposes of this paper, the set of procedures in a ConGolog
domain theory is referred to as ${\cal R}$.

We use a predicate 
$\badSituation(s)$ proposed by Reiter \cite{reiter01} to encode 
the constraints in a task network. The purpose of these constraints is 
to prune part of a search space similar to using temporal 
constraints.

To deal with partially ordered task networks, we add two new 
primitive actions $\start(P(\vec{v}))$, $\endd(P(\vec{v}))$, 
and two new fluents  $\executing(P(\vec{v}), s)$ and 
$\terminated(X, s)$, 
where $P(\vec{v})$ is a ConGolog procedure 
and $X$ is either $P(\vec{v})$ 
or an action $a \in \mathcal{A}$. 
$\executing(P(\vec{v}), s)$ states
that $P(\vec{v})$ is executing in situation $s$, 
$\terminated(X, s)$ states that $X$ has
terminated in $s$. $\executing(a, s)$ where $a \in {\cal A}$ is defined to be false. The successor state axioms for these fluents follow.  They show how the actions  $\start(P(\vec{v}))$,  $\endd(P(\vec{v}))$ 
change the truth value of these fluents:

\begin{small}\noindent \hspace*{1cm} $\executing(P(\vec{v}), do(a, s)) \equiv$   $a=\start(P(\vec{v})) \vee$ \\
\hspace*{1.7cm}  $\executing(P(\vec{v}), s) \wedge a \neq \endd(P(\vec{v}))$\\
\noindent \hspace*{1cm} $terminated(X, do(a, s)) \equiv X=a \vee$ \\
\hspace*{1.7cm}  $(X \in {\cal R} \wedge a= \endd(X)) \vee \terminated(X, s)$ \end{small}

\noindent where ${\cal R}$ is the set of ConGolog procedures in our domain.

\begin{small}
\begin{definition}[Preference-based HTN in Situation Calculus] 
\label{def:pbhtnsc}
An HTN planning problem with user preferences described as a 4-tuple $\mathcal{P}= (s_{0}, w, D, \Phi_{htn})$ 
 is encoded in situation calculus as a 5-tuple $(\atheory, \mathcal{C}, \Delta, \delta_{0}, \Phi_{sc})$ where $\atheory$ is the basic action theory, $\mathcal{C}$ is the set of ConGolog axioms,$\Delta$ is the sequence of procedure declarations for all ConGolog procedures in ${\cal R}$, $\delta_{0}$ is an encoding of the initial task network in ConGolog, and $\Phi_{sc}$ is a mapping of the preference formula $\Phi_{htn}$ in situation calculus. A plan $\vec{a}$  is a solution to the encoded preference-based HTN problem if and only if: \\
\hspace*{0.75cm} ${\cal D} \cup \mathcal{C} \models (\exists s)Do(\Delta; \delta_{0}, S_0, s) \wedge s= do(\vec a , S_{0})$
\hspace*{1.25cm}  $ \wedge~ \neg \badSituation(s) \wedge \nexists s'.[Do(\Delta; \delta_{0}, S_0, s')$ \\
\noindent \hspace*{1.25cm}  $\wedge~ \neg \badSituation(s') \wedge pref(s', s, \Phi_{sc})]$ 
\end{definition}
\end{small}

\noindent where $pref(s', s, \Phi_{sc})$ 
denotes that the situation $s'$ is preferred to situation s 
with respect to the preference formula $\Phi_{sc}$, and
{\small
$Do(\delta,S_0,do(\vec{a},S_0))$} denotes that the ConGolog program
{\small $\delta$}, starting execution in {\small $S_0$} will legally
terminate in situation {\small $do(\vec{a},S_0)$}. 
Removing all the $\start(P(\vec{v}))$ and $end(P(\vec{v}))$ actions from $\vec{a}$ to obtain $\vec{b}= (b_{1}, ..., b_{n})$, a preferred plan for the original HTN planning problem $\mathcal{P}$ is a plan $\pi=(o_{1}, ..., o_{n})$ where for all $1 \leq i \leq n$, name($o_{i}$)= $b_{i}$.

\section{HTN Preference Specification}
\label{section-htnpref}

In this section, we describe how to specify the preference formula 
$\Phi_{htn}$. Our preference language, \lph, modifies and extends 
the \lpp~qualitative preference language
proposed in \cite{biefrimcikr06} to capture HTN-specific preferences.

Our \lph~language has the ability to express 
preferences over certain 
parameterization of a task (e.g., preferring one task grounding
to another), over a certain decomposition of nonprimitive 
tasks (i.e., prefer to apply a certain method over another), and a 
soft version of the before, after, and in between constraints. A 
soft constraint is defined via a preference formula whose evaluation
determines when a plan is \emph{more preferred} than another. 
However, unlike the task network 
constraints which will prune or eliminate those plans that have not 
satisfied them, not meeting a soft constraint
simplify deems a plan to be of poorer quality.

\begin{definition}[Basic Desire Formula (BDF)]
\begin{small}
A basic desire formula is a sentence drawn from the smallest set
$\mathcal{B}$ where:\\
\hspace*{.5cm}1. If $l$ is a literal, then $l \in \mathcal{B}$ and $\final(l) \in \mathcal{B}$ \\
\hspace*{.5cm}2. If $t$ is a task, then $\occ(t) \in \mathcal{B}$\\
\hspace*{.5cm}3. If $m$ is a method, and $n=name(m)$, then  $\apply(n) \in \mathcal{B}$\\
\hspace*{.5cm}4. If $t_{1}$, and $t_{2}$ are tasks, and $l$ is a literal, then \\
\hspace*{.75cm} $\before(t_{1}, t_{2}), \holdBefore(t_{1}, l), \holdAfter(t_{1}, l),$ \\
\hspace*{.75cm} $\holdBetween(t_{1}, l, t_{2})$ are in $\mathcal{B}$.  \\
\hspace*{.5cm}5. If  $\varphi_1$ and $\varphi_{2}$ are in $\mathcal{B}$, then so are $\neg \varphi_1$, $\varphi_{1} \wedge \varphi_{2}$, 
$\varphi_{1} \vee \varphi_{2}$, \\
\hspace*{.75cm} 
$(\exists x)\varphi_1$,
$(\forall x)\varphi_1$,
\nextt($\varphi_1$),
\always ($\varphi_1$), \eventually ($\varphi_1$), \\
\hspace*{.75cm} 
and \until ($\varphi_1$, $\varphi_{2}$). 
\end{small}
\end{definition}

\noindent
{\small \final($l$)} states that the literal {\small $l$} holds in the  final state, 
{\small \occ($t$)} states that the task {\small $t$} occurs in the present state, and
{\small \nextt($\varphi_1$), \always ($\varphi_1$), \eventually ($\varphi_1$),
and \until ($\varphi_1$, $\varphi_{2}$)} are basic LTL constructs. {\small \apply($n$)} states that a method whose name is $n$ is applied to decompose a nonprimitive task. {\small \before($t_{1}, t_{2}$)} states a precedence ordering between two tasks. {\small \holdBefore($t_{1}, l$), \holdAfter($t_{1}, l$), \holdBetween($t_{1}, l, t_{2}$)} state a soft constraint over when the fluent $l$ is preferred to hold. (i.e.,  {\small \holdBefore($t_{1}, l$)} state that $l$ must be true right before the last operator descender of $t_{1}$ occurs). 
Combining $\occ(t)$ with the rest of \lph~language enables the construction
of preference statements over parameterizations of tasks. 

BDFs establish properties of different states within a plan. 
By combining BDFs using boolean and temporal connectives, we are 
able to express other properties of state. 
The following are a few examples from our travel domain\footnote{To simplify the examples many parameters have been suppressed, and we abbreviate 
\eventually(\occ($\varphi$)) by \occ$^{\prime}$, \eventually(\apply($\varphi$)) by \apply$^{\prime}$ and refer to preferences by their labels.}.

\vspace*{-.17in}
\begin{small} \begin{eqnarray} 
& &  \hspace*{-.2in} (\exists c). \occ^{\prime}(\textit{book-car}(c, Enterprise))  \label{enterprise} \\
& &  \hspace*{-.2in} \apply^{\prime} (\textit{by-car-local(SUV, Avis)}) \label{avis} \\
& &  \hspace*{-.2in} \before(\textit{arrange-trans}, \textit{arrange-acc}) \label{before} \\
& & \hspace*{-0.2in} \holdBefore(hotelReservation, \textit{arrange-trans}) \label{holdbefore} \\
& &  \hspace*{-.2in} \always( \neg (\occ^{\prime} (\textit{pay}(Mastercard) )  ) )   \label{payalways} \\
& &  \hspace*{-.2in} (\exists h,r). \occ^{\prime}(\textit{book-hotel}(h,r)) \wedge \textit{starsGE}(r, 3) \label{hotel} \\
& &  \hspace*{-.2in} (\exists c). \occ^{\prime}(\textit{book-flight}(c, \textit{Economy}, \textit{Direct}, \textit{WindowSeat}))  \nonumber \\
&& \hspace*{.2in} \wedge~ \textit{member}(c, \textit{StarAlliance}) \label{flight}
\end{eqnarray} 
\end{small}
\vspace*{-.17in}

\ref{enterprise} states that at some point the user books a car 
with Enterprise. 
\ref{avis} states that at some point, the \emph{by-car-local} method 
is applied to book an SUV from Avis.  
\ref{before} states that the \emph{arrange-trans} task occurs 
before the \emph{arrange-acc}~ task.  \ref{holdbefore} states that the 
hotel is reserved before transportation is 
arranged. \ref{payalways} states that the user never pays
by Mastercard. \ref{hotel} states that at some point the user books a hotel that has a rating of 3 or more. \ref{flight} states that at some point the user books a direct economy window-seated flight with a Star Alliance carrier.

To define a preference ordering over alternative properties of states,
\emph{Atomic Preference Formulae} (APFs) are defined.   Each alternative
comprises two components:  the property of the
state, specified by a BDF, and a \emph{value} term
which stipulates the relative strength of the preference.

\begin{definition}[Atomic Preference Formula (APF)] ~~~\\
\begin{small}
Let $\vset$ be a totally ordered set with minimal element
$\minval$ and maximal element $\maxval$. An atomic preference
formula is a formula $ \varphi_{0}[v_0] \gg \varphi_{1}[v_1] \gg
... \gg \varphi_{n}[v_n]$, where each $\varphi_{i}$ is a BDF,
each $v_i \in \vset$, $v_i < v_j$ for $i < j$, and $v_0 =
\minval$. When $n=0$, atomic preference formulae correspond to
BDFs.
\end{small}
\end{definition} 
While one could let $\vset = [0,1]$,
you could choose a strictly qualitative set
like $\{\textit{best} < \textit{good}
< \textit{indifferent} < \textit{bad} < \textit{worst}\}$ to express preferences over alternatives.

Now here are a few  APF examples from the travel domain. 

\vspace*{-.17in}
\begin{small} \begin{eqnarray} 
& & \hspace*{-.2in}  \ref{avis}[0] \gg  \apply^{\prime} (\textit{by-car-local(SUV, National)})  [0.3] \label{caratomic} \\
& & \hspace*{-.2in} \apply^{\prime} (\textit{by-car-trans}) [0] \gg   \apply^{\prime} (\textit{by-flight}) [0.4] \label{decompose} \\
& & \hspace*{-.2in} \occ^{\prime} (\textit{book-train}) [0] \gg   \occ^{\prime} (\textit{book-car}) [0.4] \label{prefertrain}
\end{eqnarray} 
\end{small}
\vspace*{-.17in}

\ref{caratomic} states that the user prefers that the \textit{by-car-local} 
method rents an SUV and that the rental car company Avis is 
preferred to {\small National}. 
\ref{decompose} states that the user prefers to decompose the 
\textit{arrange-trans} task by the method \textit{by-car-trans} rather 
than the \textit{by-flight} method.  Note that the task is implicit in
the definition of the method. \ref{prefertrain} states that the user prefers travelling by train over renting a car.

To allow the user to specify more complex preferences and to
aggregate preferences,
General Preference Formulae (GPFs) extend the language to
conditional, conjunctive, and disjunctive preferences.

\begin{definition}[General Preference Formula (GPF)] ~~~\\
\begin{small}
A formula $\Phi$ is a GPF if one of the following
holds:\\
\hspace*{.5cm}$\bullet$~ $\Phi$ is an APF\\
\hspace*{.5cm}$\bullet$~ $\Phi$ is $\gamma : \Psi$, where $\gamma$ is a BDF and
$\Psi$ is a GPF [Conditional]\\
\hspace*{.5cm}$\bullet$~ $\Phi$ is one of $ \,$ $\Psi_{0} \gand  \Psi_{1} \gand ... \gand \Psi_{n}$ [General Conjunction]\\
        \hspace*{2.0cm} or $\Psi_{0} \gor \Psi_{1} \gor ... \gor \Psi_{n}$
[General Disjunction]\\
\noindent where $n \geq 1$ and each $\Psi_{i}$ is
a GPF. 
\end{small}
\end{definition}

General conjunction (resp.general disjunction) refines the ordering defined by \begin{small}$\Psi_{0} \gand  \Psi_{1} \gand ... \gand \Psi_{n}$\end{small} (resp. {\small $\Psi_{0}|\Psi_{1}|...|\Psi_{n}$}) by sorting indistinguishable states using the lexicograping ordering.  Continuing our example:

\vspace*{-.17in}
\begin{small} \begin{eqnarray} 
& & \hspace*{-.2in} \occ(\textit{arrange-trans}) : (\exists c). \occ^{\prime}(\textit{book-car}(c, Avis)) \label{if-trans} \\
& & \hspace*{-.2in} \occ(\textit{arrange-local-trans}) : \ref{enterprise}  \label{if-local}\\
& & \hspace*{-.2in} drivable : \ref{prefertrain} [0] \gg \occ^{\prime}(\textit{book-flight}) [0.3]  \label {drive} \\
& & \hspace*{-.2in}    \ref{holdbefore} \gand \ref{hotel} \gand \ref{flight} \gand \ref{caratomic}  \gand \ref{decompose} \gand \ref{prefertrain} \gand  \ref{if-local} \gand \ref{drive} \label{all} 
\end{eqnarray} 
\end{small}
\vspace*{-.17in}

\ref{if-trans} states that if inter-city transportation is being arranged then the user prefers to rent a car from Avis. \ref{if-local} states that if local transportation is being arranged the user prefers 
Enterprise. \ref{drive} states that if the distance between the origin and the destination is drivable then the user prefers to book a train over booking a car over booking a flight. \ref{all} aggregates preferences into one formula.

Again, and only for the purpose of proving properties, we 
provide an encoding of the HTN-specific terms of \lph~in 
the situation calculus.  As such, for any preference formula $\Phi_{htn}$
there is a corresponding formula $\Phi_{sc}$ where every HTN-specific
term is replaced as follows:
each literal $l$ is mapped to a fluent or non-fluent relation in the situation
calculus, as appropriate; each primitive 
task $t$ is mapped to an action $a \in \mathcal{A}$; and each nonprimitive 
task $t$ and each method $m$ is mapped to a procedure $P(\vec{v}) \in \cal{R}$ in ConGolog.

\subsection{The Semantics}

The semantics of \lph~is achieved through assigning a weight to a situation $s$ with respect to
a GPF, $\Phi$, written $w_s(\Phi)$.  
This weight is a composition of
its constituents.  For BDFs, a situation $s$ is assigned
the value $v_{min}$ if the BDF is satisfied in $s$, 
$v_{max}$ otherwise. 
Similarly, given an APF, and a situation $s$, $s$ is assigned the weight 
of the best BDF that it satisfies within the defined APF.  
Finally GPF semantics follow the natural semantics of boolean connectives.
As such General Conjunction yields the minimum of its constituent 
GPF weights and
General Disjunction yields the maximum. 

Similar to \cite{precontr} and following \lpp, we use the notation $\varphi[s', s]$ to denote that $\varphi$ holds in the sequence of situations starting from $s'$ and terminating in $s$. Next, we will show how to interpret BDFs in the situation calculus.

If $f$ is a fluent, we will write {\small $f[s', s] = f[s']$} since fluents are represented in situation-suppressed form. 
If $r$ is a non-fluent, we will have $r[s', s] = r$ since $r$ is already a situation calculus formula.  Furthermore, we will write $\final(f)[s', s] = f[s]$ since $\final(f)$ means that the fluent $f$ must hold in the final situation. 

The BDF $\occ(X)$ states the occurrence of $X$ which can be either an action or a procedure.  written as: \\
\vspace*{.10cm}
\hspace*{0.75cm} \begin{small}$\occ(X)[s', s] =  \left \{\begin{array}{l} do(X, s') \sqsubseteq s \tab \tab \mbox{if X} \in \mathcal{A} \\
do(start(X), s') \sqsubseteq s \;  \: \; \mbox{if X} \in \mathcal{R}
\end{array}
\right.$ 
\end{small} 

\noindent The BDF  {\small $\apply(P(\vec{v}))$} will be interpreted as follows: \\
\vspace*{.10cm}
\hspace*{0.75cm} \begin{small} $\apply(P(\vec{v}))[s', s] = do(start(P(\vec{v})), s') \sqsubseteq s$\end{small}

\noindent Boolean connectives and quantifiers are already part of the situation calculus and require no further explanation here.  The LTL constructs are interpreted in the same way as in \cite{precontr}. We interpret the rest of the connectives as follows \footnote{We use the following abbreviations: \\
\hspace*{0.5cm} $(\exists s_1: s' \sqsubseteq  s_1 \sqsubseteq  s) \Phi = (\exists s_1) \lbrace s' \sqsubseteq  s_1 \wedge s_1 \sqsubseteq s \wedge \Phi \rbrace$ \\
\hspace*{0.5cm} $(\forall s_1: s' \sqsubseteq  s_1 \sqsubseteq  s) \Phi = (\forall s_1) \lbrace[s' \sqsubseteq  s_1 \wedge s_1 \sqsubseteq s] \subset \Phi \rbrace$}.

\noindent \begin{small}\hspace*{.85cm}$\before(X_1, X_2)[s',s] = (\exists s_1, s_2:s' \sqsubseteq s_1 \sqsubseteq s_2 \sqsubseteq s)$ \\ 
\hspace*{.7cm}$\bigtab \{ \terminated(X_1)[s_1] \wedge \neg  \executing(X_2)[s_1] $\\
\hspace*{.7cm}$\bigtab \wedge~ \neg \terminated(X_2)[s_1] \wedge \occ(X_2)[s_2, s] \}$\\
\hspace*{.85cm}$\holdBefore(X, f)[s',s] = (\exists s_1: s' \sqsubseteq s_1 \sqsubseteq s)$\\
\hspace*{.7cm}$\bigtab \{ f[s_1] \wedge \occ(X)[s_1, s] \}$\\
\hspace*{.85cm}$\holdAfter(X, f)[s',s]= (\exists s_1: s' \sqsubseteq s_1 \sqsubseteq s) $\\
\hspace*{.7cm}$\bigtab \{ terminated(X)[s_1] \wedge f[s_1]  \}$ \\ 
\hspace*{0.85cm}$\holdBetween(X_1, f, X_2)[s', s] = $\\
\hspace*{.7cm}$\bigtab (\exists s_1, s_2:s' \sqsubseteq s_1 \sqsubseteq s_2 \sqsubseteq s) $\\
\hspace*{.7cm}$\bigtab \{ \terminated(X_1)[s_1] \wedge \neg  \executing(X_2)[s_1] $\\
\hspace*{.7cm}$\bigtab \wedge~ \neg \terminated(X_2)[s_1] \wedge \occ(X_2)[s_2, s] \}$ \\
\hspace*{.7cm}$\bigtab \wedge~  (\forall s_i: s_1 \sqsubseteq s_i  \sqsubseteq s_2) f[s_i] $ \end{small}

\noindent From here, the semantics follows that of \lpp.

\begin{small}
\begin{definition}[Basic Desire Satisfaction]\label{gbdsat}
Let $\atheory$ be an action theory, and let $s'$ and $s$ be  situations such
that
$s' \sqsubseteq s$. The situations 
beginning in $s'$ and terminating in $s$ satisfy $\varphi$ just in the case 
that
{\small $\atheory \models \varphi[s',s]$.}
We define $w_{s', s}(\varphi)$ to be the weight of the
situations originating in $s'$ and ending in $s$ wrt BDF 
$\varphi$.  $w_{s', s}(\varphi)=\minval$ if $\varphi$ is satisfied,
otherwise $w_{s',s}(\varphi)=\maxval$. 
\end{definition}  
\end{small}

Note that for readability we are going to drop $s'$ from the index, i.e.,  $w_{s}(\varphi) = w_{s',s}(\varphi)$ in the special case of $s' = S_0$.

\begin{small}
\begin{definition}[Atomic Preference Satisfaction]
Let $s$ be a situation and 
$\Phi = \varphi_{0}[v_0]  \gg \varphi_{1}[v_1] \gg ... \gg \varphi_{n}[v_n]$ be an atomic  
preference formula. Then $w_s(\Phi)= v_i$ if $i =\min_j \{\atheory \models \varphi_j[S_0,s]\}$, and $w_s(\Phi)= \maxval$ if no such $i$ exists. 
\end{definition} 
\end{small}

\begin{small}
\begin{definition}[General Preference Satisfaction]
\label{def-GPS}
Let $s$ be a situation and $\Phi$ be a general preference
formula. Then $w_s(\Phi)$ is defined as follows:
\begin{small}

\noindent \hspace*{.5cm}$\bullet$~ $w_s(\varphi_{0}  \gg \varphi_{1} \gg ... \gg \varphi_{n})$ is 
defined above \\
\hspace*{.5cm}$\bullet$~ $w_s(\gamma : \Psi)= \left \{ 
		\begin{array}{lcl}
                     \minval & & \mbox{if} \: w_s(\gamma)=\maxval \\                                     					   w_s(\Psi) & & \mbox{otherwise} \\   
		\end{array} \right.$    \\
\hspace*{.5cm}$\bullet$~ \mbox{$w_s(\Psi_{0} \gand \Psi_{1} \gand ... \gand 
      \Psi_{n})= \max \{w_s(\Psi_{i}) : 1 \leq i \leq n\} $}  \\
\hspace*{.5cm}$\bullet$~ $w_s(\Psi_{0} \gor \Psi_{1} \gor ... \gor \Psi_{n})= 
     \min \{w_s(\Psi_{i}) : 1 \leq i \leq n\} $  

\end{small} 
\end{definition} 
\end{small}

The following definition dictates how to compare two situations 
(and thus two plans) with respect to a GPF.  This preference relation 
{\small $pref$} is used to compare HTN plans in 
Definition \ref{def:pbhtnsc} and provides the semantics for 
\emph{more preferred} in Definition \ref{def:pbhtnp}.

\begin{small}
\begin{definition}[Preferred Situations]
\label{ordering}
A situation $s_1$ is at least as preferred as a situation $s_2$
with respect to a GPF $\Phi$, written $pref(s_1,s_2,\Phi)$
if $w_{s_1}(\Phi) \leq w_{s_2}(\Phi)$.
\end{definition}
\end{small}

\section{Computing Preferred Plan}
\label{heuristic}

To compute a preferred plan, we proposed a heuristic-search, 
forwarding-chaining planner that searchs for the \textit{most preferred} 
terminating state that satisfies the HTN planning problem. 
The search is guided by an admissible evaluation function that evaluates 
partial plans with respect to preference satisfaction. 
We
use \textit{progression} to evaluate the preference formula 
satisfaction over partial plans. 

\subsection{Progression}

Given a situation and a temporal formula, progression evaluates it 
with respect to the state of a situation to generate a new formula 
representing  those aspects of the formula that remain to be satisfied. In this section, we define the progression of the constructs we added/modified from \lpp~and show that progression preserves the semantics  
of preference formulae. To define the progression, similar to \cite{biefrimcikr06} we add the 
propositional constants $\true$ and $\false$ to both the situation calculus 
and to our set of BDFs, where $\atheory \vDash$ $\true$ 
and $\atheory \nvDash \false$ for every action 
theory $\atheory$. We also add the BDF {\small $\occnext(X)$}, 
and {\small $\applynext(P(\vec{v}))$} to capture the progression 
of {\small $\occ(X)$} and {\small $\apply(P(\vec{v}))$}. Below we show the progression of the added constructs. 
\begin{small}
\begin{definition}[Progression] Let $s$ be a situation, and let $\varphi$ be a BDF.
The progression of $\varphi$ through $s$, written $\pr(\varphi)$, is given by: \\
$\bullet ~$If $\varphi$=\occ(\textit{X}) then  \\
\hspace*{0.2cm} $\pr(\varphi)$ = \textbf{occNext}$($\textit{X}$)$ $\wedge \eventually(terminated($\textit{X}$))$ \\
$\bullet$ If $\varphi= $ \textbf{occNext}($X$) , then \\
\hspace*{0.04cm} $\left \{
\begin{array}{lll}  
\true & & \mbox{if} \;  \;X \in {\cal {A}} \wedge \atheory \models \exists s'. s=do(X, s') \\
\true & & \mbox{if} \; \; X \in {\cal {R}} \wedge \atheory \models \exists s'. s=do(start(X), s') \\
\false & &  \mbox{otherwise}
\end{array}
\right.$\\
$\bullet$ If $\varphi= \apply(P(\vec{v})), \mbox{then}$ \\
\hspace*{0.2cm} $\pr(\varphi) = \mbox{\textbf{applyNext}} (P(\vec{v}))$ $\wedge \eventually(terminated(P(\vec{v})))$ \\
$\bullet$ If $\varphi= \mbox{\textbf{applyNext}} (P(\vec{v}))$ , then \\
\hspace*{0.2cm} $\pr(\varphi) = \left \{
\begin{array}{lll} 
\true & & \mbox{if} \; \; \atheory \models \exists s'.s=do(start(P(\vec{v})), s') \\
\false & & \mbox{otherwise}
\end{array}
\right.$\\
$\bullet$ If $\varphi = \before(X_{1}, X_{2})$, $\holdBefore(X, f)$, $\holdAfter(X, f)$, \\
\hspace*{0.5cm}    or $\holdBetween(X_{1}, f, X{2})$, then \\
\hspace*{1cm} $\pr(\varphi) = \left \{
\begin{array}{lll} 
\true & & \mbox{if} \; \; w_{s}(\varphi) = v_{min} \\
\false & & \mbox{otherwise}
\end{array}
\right.$
\end{definition}
\end{small}

To see how the other constructs are progressed please refer to \cite{biefrimcikr06}.

\subsection{Admissible Evaluation Function}
\def\osats{[s',s]^{opt}}
\def\psats{[s',s]^{pess}}
\def\opsats{[s',s]^{opt/pess}}

In this section, we describe an admissible evaluation function 
using the notion of 
{\em optimistic} and {\em pessimistic} weights
that provide a bound on the best
and worst weights of any successor situation with respect to a GPF
$\Phi$. Optimistic (resp. pessimistic) weights, {\small $w_{s}^{opt}(\Phi)$}
(resp. {\small $w_{s}^{pess}(\Phi)$}) are defined based 
on optimistic (resp. pessimistic) satisfaction of BDFs. 
Optimistic satisfaction ($\varphi[s',s]^{opt}$) assumes that any parts of
the BDF not yet falsified will eventually be satisfied.
Pessimistic satisfaction ($\varphi[s',s]^{pess}$) assumes the
opposite.
The following definitions highlight the key 
differences between this work and the definitions in \cite{biefrimcikr06}. \\
\begin{small}
\hspace*{0cm} $\occ(X)  \osats  \isdef   \left \{\begin{array}{l} do(X, s') \sqsubseteq s \vee s'=s \tab \; \; \; \; \;  \mbox{if X} \in \mathcal{A} \\
do(start(X), s') \sqsubseteq s \vee s'=s \;   \mbox{if X} \in \mathcal{R}
\end{array}
\right.$  \\
\hspace*{0cm} $\occ(X)  \psats \isdef  \left \{\begin{array}{l} do(X, s') \sqsubseteq s \tab \; \; \; \; \; \; \mbox{if X} \in \mathcal{A} \\
do(start(X), s') \sqsubseteq s \;  \; \mbox{if X} \in \mathcal{R}
\end{array}
\right.$  \\
\hspace*{0cm} $\apply(P(\vec{v})) \osats  \isdef  do(start(P(\vec{v})), s') \sqsubseteq s \vee s'=s$ \\
\hspace*{0cm} $\apply(P(\vec{v})) \psats \isdef  do(start(P(\vec{v})), s') \sqsubseteq s$ \\
\hspace*{0cm} If $\varphi = \before(X_{1}, X_{2}), \holdBefore(X,f), \holdAfter(X, f)$ \\
\hspace*{1cm} $\holdBetween(X_{1}, f, X_{2})$, then \\
\hspace*{1cm} $\varphi \osats  \isdef \varphi \psats  \isdef w_{s', s}(\varphi)$ 
\end{small}

\begin{small}\begin{thm}
\label{thm-optpess}
Let {\small $s_n = do([a_1,...,a_n],S_0), n \geq 0$} be a collection of
situations, {\small $\varphi$} be a BDF, {\small $\Phi$} 
a general preference formula, and $w_{s}^{opt}(\Phi)$, $w_{s}^{pess}(\Phi)$ be the optimistic and pessimistic weights of $\Phi$ with respect to $s$. Then for any ${\small 0 \leq i \leq j \leq k
\leq n}$, \\
{\fs{8pt}
\noindent 1. ${\small \atheory \models \varphi[s_i]^{pess}
\Rightarrow \atheory \models \varphi[s_j]}$, ${\small \atheory \not\models \varphi[s_i]^{opt} \Rightarrow
\atheory \not\models \varphi[s_j]}$, \\
\noindent 2. ${\small \left( w_{s_i}^{opt}(\Phi) = 
w_{s_i}^{pess}(\Phi) \right) \Rightarrow}$ ${\small w_{s_j}(\Phi) = w_{s_i}^{opt}(\Phi) = w_{s_i}^{pess}(\Phi)}$, \\
\noindent 3. \begin{scriptsize}$w_{s_i}^{opt}(\Phi) \leq w_{s_j}^{opt}(\Phi) 
\leq w_{s_k}(\Phi)$, $w_{s_i}^{pess}(\Phi) \geq w_{s_j}^{pess}(\Phi) \geq w_{s_k}(\Phi)$\end{scriptsize}
}

\end{thm}\end{small}

\noindent Theorem \ref{thm-optpess} states that the optimistic weight
is non-decreasing and never over-estimates the real weight. Thus,
$f_\Phi$ is admissible and when used in best-first search, the
search is optimal.

\begin{small}\begin{definition}[Evaluation function]
Let $s=do(\vec{a},S_0)$ be a situation and let $\Phi$ be a
general preference formula. Then $f_\Phi(s) \isdef w_s(\Phi)$ if $\vec{a}$ is a plan, otherwise $f_\Phi(s) \isdef w_{s}^{opt}(\Phi)$.
\end{definition}\end{small}


\section{Implementation and Results}
\label{section-results}

\renewcommand{\arraystretch}{1.2}

In this section, we describe our best-first search, ordered-task-decomposition 
planner. 
Figure \ref{fig:htnpref} outlines the algorithm. \htnpref~takes as input $\mathcal{P} = (s_{0}, w, D, pref)$ where $s_{0}$ 
is the initial state, $w$ the initial task network, $D$ is the HTN 
planning domain, and \textit{pref} the general preference formula, 
and returns a sequence of ground primitive operators, i.e. a plan, and 
the weight of that plan.

The  {\it frontier} is a list of nodes of the form  {\small [\textit{optW}, 
\textit{pessW}, 
$w$, 
\textit{partialP}, \textit{s}, \textit{pref}]}, sorted
by optimistic weight, pessimistic weight, and then by plan length. 
The frontier is initialized to the initial task network $w$, the empty partial plan, its {\small {\it optW}, {\it pessW}, 
and {\it pref}}~corresponding to the progression and  evaluation of the input preference formula in the initial state.

On each iteration of the \textbf{while} loop, \htnpref~removes the first node
from the frontier and places it in {\it current}. If $w$ is empty (i.e., $U$ is an empty set), the situation associated with this node is a terminating situation. 
Then  \htnpref~returns {\it current}'s partial plan and weight.
Otherwise, it calls the function \expand~with {\it current's} node as input.

\expand~returns a new list of nodes that need to be added to the frontier. The new nodes are sorted by {\small \textit{optW}, \textit{pessW}}, 
and  merged with the remainder of the frontier. 
If {\small \textit{w}} is {\small \textit{nil}}~ then the frontier is left as is.
Otherwise, it generates a
new set of nodes of the form {\small [\textit{optW}, \textit{pessW}, \textit{newW}, 
\textit{newPartialP}, \textit{newS}, \textit{newProgPref}]}, 
one for each legal ground operator that can be reached by performing $w$ using a partial-order forward decomposition procedure (PFD) \cite{htn04}. Currently \htnpref~uses \shop2~\cite{shop2HTN} as its PFD.  Hence, the current
implementation of \htnpref~is an implementation of \shop2~with user 
preferences.
For each primitive task leading to terminating states,
\expand~generates a node of the same form but with \textit{optW} 
and \textit{pessW} replaced by the actual weight. 
If we reach the empty frontier, we return the empty plan. 
 
\begin{figure}[t]

\centering
\begin{boxedminipage}{80mm}
\begin{small}
{\fs{9pt}
\noindent {\htnpref}($s_{0}$, \textit{w}, \textit{D}, \textit{pref})  \\
\noindent \textit{frontier} $\leftarrow$ 
         \initfront($s_{0}$, \textit{w},  \textit{pref})\\
\textbf{while} \textit{frontier} $\neq \emptyset$ \\
\tab \textit{current} $\leftarrow$ \remfirst(\textit{frontier})\\
\tab \% establishes  values of
   \textit{w}, \textit{partialP}, \textit{s}, \textit{progPref} \\
\tab \textbf{if} \textit{w}= $\emptyset$ and \textit{optW}=\textit{pessW} then \textbf{return} \textit{partialP}, \textit{optW} \\
\tab \textit{neighbours} $\leftarrow$ 
\expand(\textit{w}, \textit{D}, \textit{partialP}, \textit{s}, \textit{progPref})\\
\tab \textit{frontier} $\leftarrow$ \sortnmerge(\textit{neighbours},
     \textit{frontier})\\
 \textbf{return} [], $\infty$

}
\end{small}
 \end{boxedminipage}
  \caption{A sketch of the \htnpref~algorithm.}
  \label{fig:htnpref}

\end{figure}

\begin{thm}[Soundness and Optimality]   $\,$ \\
\label{thm: correctness}
\begin{small}
Let $\cal{P}$=$(s_{0},w, D, \Phi)$ be a HTN planning problem with user preferences.  Let $\pi$ be the plan returned by \htnpref~from input $\cal{P}$.  Then $\pi$ is a solution to the preference based HTN problem $\mathcal{P}$
\end{small}
\end{thm}
\noindent
{\it Proof sketch:}  We prove that the algorithm terminates appealing to the fact that the PFD procedure is sound and complete. We prove that the returned plan is optimal, by exploiting the correctness of progression of preference formula, and admissibility of our evaluation function.

\subsection{Experiments}

We implemented our preference-based HTN planner, \htnpref, on top of the LISP 
implementation of \shop2~\cite{shop2HTN}.  All experiments were run on a 
Pentium 4 HT, 3GHZ CPU, and 1 GB RAM, with a time limit of 900 seconds. 
Since the optimality of
\htnpref-generated plans was established in Theorem \ref{thm: correctness},
our objective was to evaluate the effectiveness of our heuristics in 
guiding search towards the optimal plan, and to establish benchmarks for 
future study, since none currently exist.

We tested \htnpref~with ZenoTravel and Logistics domains,
which were adapted from the International Planning Competition
(IPC). 
The ZenoTravel domain involves transporting people on aircrafts that
can fly at two alternative speeds between locations. 
The Logistics domain involves transporting packages to different destinations using trucks for delivery within cities and planes for
between cities.

In order to evaluate the effectiveness of \htnpref~it would have been appealing to evaluate our planner 
with a preference-based planner that also makes use of procedural control 
knowledge. But since no comparable planner exists,  
and it would not have been fair to compare \htnpref~with a 
preference-based planner that does not use control knowledge, 
we compared \htnpref~with 
\shop2, using a brute-force technique for \shop2~to determine the
optimal plan. In particular, as is often done with Markov Decision Processes,
\shop2~generated all plans that satisfied the HTN specification and then
evaluated each to find the optimal plan.  Note that the times reported
for \shop2 do not actually include the time for posthoc preference 
evaluation, so they are lower bounds on the time to compute the optimal plan.

\begin{figure}[t!]
  \centering

	{\fs{8pt}
  \subfigure[  ZenoTravel domain  ]{
    \label{labelname 1}
    
\begin{tabular}{|l||c|c|c||c|c|c|c|}

\hline

\textbf{P} & \multicolumn{3}{|c||} {\textbf{SHOP2}} & \multicolumn{4}{|c|} {\textbf{HTNPREF}}  \\

\cline{2-8}

 \# & \# \textbf{Plan} &\NE & \textbf{Time}  &  \NE & \NC &\textbf{Time} &  \PL \\

\hline

1	& 12	&172&	0.54&	79&	89&	1.71&	23 \\
\hline
2	&19	&224&	2.41&	72&	78&	2.2&	30\\
\hline
3	&155	&1629&	14.47&	160&	188&	5.71&	30\\

\hline
4	&204	&2287&	19.58&	53&	59	&0.84&	29\\
\hline
5	&230	&2235&	9.13&	362&	414&	7.75&	24\\
\hline
6	&230	&2235&	9.13&	77&	24&	1.67&	24\\
\hline
7	&485	&6332&	64.24&	241&	277&	13.58&	39\\
\hline
8	&487	&6227&	109.9&	122&	125&	13.8&	46\\
\hline
9	&720	&6725&	45.62&	212&	251&	7.96&	32\\
\hline
10	&4491	&45612&	492.1&	2154&	2923&	128.1&	36\\

\hline
11	&$>$1522	&$>$16K&	$>$900&	145&	155&	11.34&	58\\
\hline
12	&$>$2156	&$>$24K&	$>$900&	1680&	1690&	238.1&	50\\

\hline

\end{tabular}

	} 
  
  \subfigure[ Logistic domain
]{
    \label{labelname 21} 

 \begin{tabular}{|l||c|c|c||c|c|c|c|}

\hline

\textbf{P} & \multicolumn{3}{|c||} {\textbf{SHOP2}} & \multicolumn{4}{|c|} {\textbf{HTNPREF}}  \\

\cline{2-8}

 \# & \# \textbf{Plan} &\NE & \textbf{Time}  &  \NE & \NC &\textbf{Time} &  \PL \\
 
\hline

1	&	8&109&	0.28&	32&	34&	0.44&	28 \\
\hline
2	&	90&540&	1.01&	20&	25&	0.24&	13\\
\hline
3	&	92&497&	0.41&	18&	20&	0.16&	14\\
\hline
4	&	808&4597&	6.01&	302&	405&	3.47&	19\\
\hline
5	&	920&4310&	5.22&	74&	94&	1.01&	15\\
\hline

6	&	1260&6320&	6.58&	131&	173&	1.48&	15\\
\hline
7	&	2178&15104&	26.18&	28&	33&	0.39&	21\\
\hline
8	&	2520&14728&	20.07&	30&	41&	0.56&	17\\
\hline

9	&$>$35K	&$>$236K	&$>$900	&38	&49	&0.65	&25 \\
\hline
10	&$>$39K	&$>$153K	&$>$900	&905	&1246	&22.0	&21 \\
\hline
11	&$>$40K	&$>$156K	&$>$900	&1K	&1438	&20.1	&20\\
\hline
12	&$>$42K	&$>$230K	&$>$900	&452	&619	&7.88	&23\\

\hline
\end{tabular}

}

	}

  \caption{{\fs{9pt}Our criteria for comparisons are number of Nodes Expanded (NE), number of applied operators; number of Nodes Considered (NC), the number of nodes that were added to the frontier, and time measured in seconds. Note NC is  equal to NE for \shop2. PL is the Plan Length and \# Plan is the total number of plans.}}
  \label{fig:allfig}
\end{figure}

Figure \ref{fig:allfig} reports our experimental results for ZenoTravel
and the Logistics domain.  
The problems varied in preference difficulty and are shown 
in the order of difficulty with respect to number of possible plans (\# Plan)
that satisfy the HTN control.

The results show that, in all but the first case of each domain,
\shop2~required more time to find
the optimal plan, and expanded more nodes. In particular note that in
problems 11 and 12 \shop2~ran out of time
(900 seconds) while \htnpref~found the optimal plan well within the time limit.
Also note that 
\htnpref~expands far fewer nodes in comparison to \shop2, illustrating
the effectiveness of our evaluation function in guiding search.

\section{Summary and Related Work}
\label{section-relatedwork}

In this paper, we addressed the problem of generating preferred plans
by combining the procedural control knowledge of
HTNs with rich qualitative
user preferences.  The most significant contributions of this paper
include:
\lph, a rich HTN-tailored preference specification language, developed
as an extension of a previously existing language;
an approach to (preference-based) HTN planning based
on forward-chaining heuristic search, that exploits progression to
evaluate the satisfaction of preferences during planning;
a sound and optimal implementation of an ordered-task-decomposition
preference-based HTN planner; and
leveraging previous research, an encoding of HTN planning
with preferences in the situation calculus, that enabled us to prove our
theoretical results. While the implementation we present here exploits
\shop2, the language and techniques proposed are relevant to a broad
range of HTN planners.

In previous work, we addressed the problem of integrating user preferences
into Web service composition \cite{sohProMcI}. 
To that end, we developed a Golog-based composition engine
that also exploits heuristic search.  It similarly
uses an optimistic heuristic. 
The
language used in that work was \lpp~and had no Web-service or
Golog-specific extensions for complex actions.  This paper's HTN-tailored
language and HTN-based planner are significantly different.

Preference-based planning has been the subject of much interest in the
last few years, spurred on by an International Planning Competition (IPC)
track on this subject.
A number of planners were developed, all based on the the competition's 
PDDL3 language \cite{pddl3}.
Our work is distinguished in that it exploits \emph{procedural} 
(action-centric) domain control knowledge in the form of an HTN, and
action-centric and state-centric preferences in the form of \lph.
In contrast, the preferences and domain control in PDDL3 and its variants are 
strictly state-centric.  
Further, \lph~is {\it qualitative} whereas PDDL3 is quantitative,
appealing to a numeric objective function.
We contend that qualitative, action- or task-centric preferences 
are often more compelling and easier to elicit that their PDDL3 counterparts.

While no other HTN planner can perform true
preference-based planning, 
\shop2~\cite{shop2HTN} and \enquirer~\cite{Kuter2004} handle some 
simple user constraints. In particular the order of methods and
sorted preconditions in a 
domain description specifies a user preference over which method is 
more preferred to decompose a task. Hence users may write 
different versions of a domain description to specify simple
preferences. However, unlike \htnpref~the user constraints are treated as hard 
constraints and (partial) plans that do not meet these constraints 
will be pruned from the search space. 
Further, there is no way to handle temporally extended hard or soft 
constraints in \shop2.  We used progression in our approach to planning
precisely to deal with these interesting preferences.
Were we limiting the expressive power of preferences to 
\shop2-like method ordering, we would have created a different planner. 
Interestingly, \shop2~ method ordering can still be exploited in our approach,
but requires a mechanism that is beyond the scope of this paper.

Finally, the \aspen~planner \cite{aspen} performs a simple form of 
preference-based planning, focused mainly on preferences over resources
and with far less expressivity than \lph.  Nevertheless, \aspen~has
the ability to plan with HTN-like task decomposition, and as such,
this work is related in spirit, though not in approach to our work. 

\vspace*{.05in}
\noindent
{\bf Acknowledgements:}  We gratefully acknowledge funding from
the Natural Sciences and Engineering Research Council of Canada (NSERC)
and the Ontario Ministry of Research and Innovation Early Researcher Award.

\bibliographystyle{aaai} 

 \providecommand{\Proceedings}{Proceedings }
  \providecommand{\International}{International }
  \providecommand{\longshort}[2]{#1 (#2)}
  \providecommand{\longshortnopar}[2]{#1}

\end{document}